\documentclass{article}

\usepackage{arxiv}

\usepackage[utf8]{inputenc} % allow utf-8 input
\usepackage[T1]{fontenc}    % use 8-bit T1 fonts
\usepackage{hyperref}       % hyperlinks
\usepackage{url}            % simple URL typesetting
\usepackage{booktabs}       % professional-quality tables
\usepackage{amsfonts}       % blackboard math symbols
\usepackage{nicefrac}       % compact symbols for 1/2, etc.
\usepackage{microtype}      % microtypography
\usepackage{lipsum}
\usepackage{graphicx}
\usepackage{amsmath}
\usepackage{multicol}
\usepackage[linesnumbered,ruled,vlined]{algorithm2e} %By DG
\usepackage{multirow} % by DG

\SetKwInput{KwInput}{Input}   % by DG             % Set the Input
\SetKwInput{KwOutput}{Output}   % by DG           % set the Output

\usepackage[ruled,vlined]{algorithm2e}
\title{Affect Expression Behaviour Analysis In The Wild Using Consensual Collaborative Training}

\author{
  Darshan Gera \\
  \texttt{darshangera@sssihl.edu.in} \\
   \And
S Balasubramanian \\
  \texttt{sbalasubramanian@sssihl.edu.in} \\
}

\begin{document}
\maketitle
\begin{abstract}
Facial expression recognition (FER) in the wild is crucial for building reliable human-computer interactive systems. However, annotations of large scale datasets in FER has been a key challenge as these datasets suffer from noise due to various factors like crowd sourcing, subjectivity of annotators, poor quality of images, automatic labelling based on key word search etc. Such noisy annotations impede the performance of FER due to the memorization ability of deep networks. During early learning stage, deep networks fit on clean data. Then, eventually, they start overfitting on noisy labels due to their memorization ability, which limits FER performance. This report presents Consensual Collaborative Training (CCT) framework used in our submission to expression recognition track of the Affective Behaviour Analysis in-the-wild (ABAW) 2021 competition. CCT co-trains three networks jointly using a convex combination of supervision loss and consistency loss, without making any assumption about the noise distribution. A dynamic transition mechanism is used to move from supervision loss in early learning to consistency loss for consensus of predictions among networks in the later stage. Co-training reduces overall error, and consistency loss prevents overfitting to noisy samples. The performance of the model is validated on challenging Aff-Wild2 dataset for categorical expression classification. Our code is made publicly available\footnote{https://github.com/1980x/ABAW2021DMACS}.
\end{abstract}

% keywords can be removed
\keywords{Facial Expression Recognition \and Collaborative training \and Aff-Wild2 \and Noisy annotation }

%\begin{multicols}{2}

\section{Introduction}
FER is an active research area in human-computer interactive systems as expressions convey important cue about emotional affect state of individuals.  Traditional FER systems were built using facial images collected in-a-lab like environment. These methods fail to perform  well under natural and un-controlled conditions. To tackle these challenges, large scale datasets like AffectNet \cite{12} captured in-the-wild have been developed. However, collecting a large-scale dataset with accurate annotations is usually impractical. Large scale FER datasets suffer from noisy annotations due to i) ambiguity in expressions, ii) poor quality of images, iii) automatic annotations obtained by querying web using keywords, iv) prototypic expressions vary across cultures, situations, and across individuals under same situation  and v) subjectivity of annotators. Training with incorrect labels results in overfitting on wrong label images due to unique memorization ability of deep networks \cite{3_arpit2017closer, 5_45820} and model may learn wrong features which results in the degradation of performance. So, it is important to eliminate the influence of noisy samples during training.

Kollias et al. \cite{ABAW2} as a part of Affective Behavior Analysis in-the-wild (ABAW) 2021 competition have provided benchmark dataset Aff-Wild2 \cite{kollias2018aff, kollias2019expression, kollias2018multi, kollias2020analysing, kollias2021distribution, kollias2021affect, kollias2019face}  consisting of in-the-wild 542 videos with 2,786,201 frames collected from YouTube. This dataset is an extension of Aff-Wild \cite{zafeiriou2017aff, kollias2017recognition, kollias2019deep} dataset. Aff-Wild2 dataset is annotated for 3-different tasks: i) valence and arousal estimation (2-D continuous dimensional model \cite{russell1980circumplex}, ii) seven basic emotions \cite{ekman1992argument} of happiness, neutral, anger, sad, surprise, disgust and fear (categorical classification) and iii) facial action unit recognition based on Facial Action Unit Coding System \cite{ekman2002facial} (multi-label classification). Since Aff-Wild2 is of very large scale, so it is unlikely that it does not have noisy annotations. Based on our recently proposed Consensual Collaborative Training (CCT) FER framework in the presence of noisy annotations \cite{gera_cct}, we experiment and validate our model on Aff-Wild2 dataset for expression classification track in ABAW 2021.
\section{Method}
In this section, we briefly present our CCT framework proposed in \cite{gera_cct} for FER under noisy annotations.

\begin{figure*}
\centerline{\includegraphics[width=1\columnwidth]{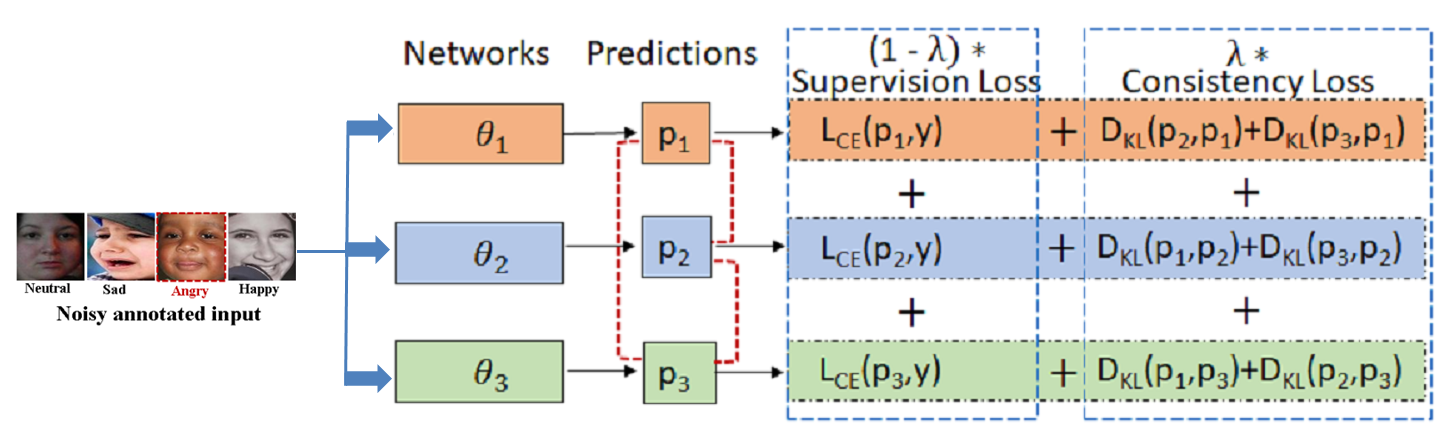}}
\caption{CCT involves training three networks $\theta_1, \theta_2 \text{ and } \theta_3$ jointly using a convex combination of supervision loss and consistency loss. Consensus is built by aligning the posterior distributions (shown as dotted red curves between $p_{1},p_{2}$ and $p_{3}$) using consistency loss. Dynamic weighting factor ($\lambda$) that balances both the losses is described in section \ref{DB}.}
\label{CCTframework}
\end{figure*}

\label{CCT}
\subsection{Overview}The CCT method follows the principle of consensus based collaborative training called Co-Training \cite{28_blum1998combining}. It uses three networks to learn robust facial expression features in the presence of noisy annotations.  Inspired by \cite{10_wei2020combating,11_sarfraz2021noisy,29_dutt2020coupled}, we use three networks with identical architecture, but different initialization, trained jointly using a convex combination of supervision loss and consistency loss. Different initialization promote different learning paths for the networks, though they have same architecture. This subsequently reduces the overall error by averaging out individual errors due to the diversity of predictions and errors. In the initial phase of training, networks are trained using supervision loss. This ensures that clean data is effectively utilized during training since DNNs fit clean labels initially \cite{3_arpit2017closer, 5_45820}. Further, to avoid eventual memorization of noisy labels by individual DNNs \cite{3_arpit2017closer, 5_45820}, gradually, as the training progresses, more focus is laid on consensus building using consistency loss between predictions of different networks. Building consensus between networks of different capabilities ensures that no network by itself can decide to overfit on noisy labels. Further, it also promotes hard instance learning during training because the networks are not restricted to update based on only low loss samples, and further that, as the training progresses, they must agree.  The trade-off between supervision and consistency loss is dynamically balanced using a Gaussian-like ramp-up function \cite{11_sarfraz2021noisy}. Further, the proposed CCT does not require noise distribution information, and it is also architecture independent. Figure \ref{CCTframework} delineates the proposed CCT framework and Algorithm \ref{CCTtraing_algorithm} enumerates the pseudo-code for CCT training. CCT implicitly avoids memorization because it has three networks collaborating with each other through the consistency loss.

\subsection{Problem formulation}
Let $D=\{(x_{i}, y_{i})\}_{i=1}^{N}$ be the dataset of N samples where $x_{i}$ is the $i^{th}$ input facial image with expression label $y_{i}\in\{1,2,...,C\}$, C denoting the number of expressions. We formulate CCT as a consensual collaborative learning between three networks parametrized by $\{\theta_{j}\}_{j=1}^{3}$. The learning is achieved by minimizing the loss $L$ given by:
\begin{equation} \label{loss}
L = (1-\lambda) * L_{sup} + \lambda * L_{cons} 
\end{equation}
where $L_{sup}$, $L_{cons}$ and $\lambda$ are described below.

\subsubsection{Supervision loss} Cross-entropy ($CE$) loss is used as supervision loss to minimize the error between predictions and labels. Let $p_j$ denote the prediction probability distribution of network $j$. Then,
\begin{equation} \label{sup_loss}
 L_{sup} = \sum_{j=1}^{3}  L_{CE}^{(j)}(p_{j},y)
\end{equation}
where y is the groundtruth vector.

\subsubsection{Consistency loss} We use Kullback-Leibler divergence ($D_{KL}$) to bring consensus among predictions of different networks by aligning their  probability distributions.
\begin{equation} \label{cons_loss}
 L_{cons}  = \sum_{j=1}^{3} \mathop{\sum_{k=1}^{3}}_{k\neq j} D_{KL}^{(j)}(p_{k}||p_{j})
\end{equation}

\subsubsection{Dynamic balancing} 
\label{DB}
The dynamic trade-off factor between $L_{sup}$ and $L_{cons}$ is computed as in \cite{11_sarfraz2021noisy}. Specifically, 
\begin{figure*}
\centerline{\includegraphics[width=0.6\columnwidth]{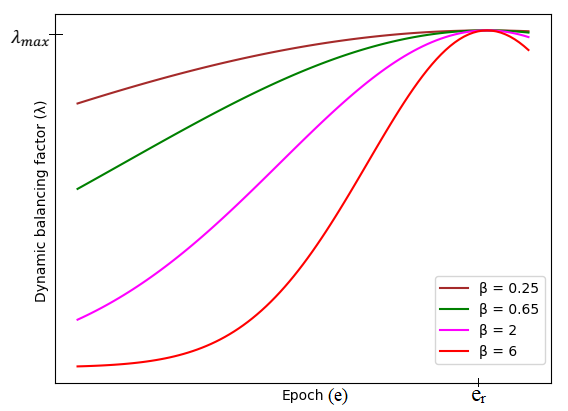}}
\caption{Dynamic balancing curves for different values of $\beta$ }
\label{dynamic_balancing_factor}
\end{figure*}
\begin{equation} \label{rampup}
 \lambda = \lambda_{max} * \exp^{-\beta(1 - \frac{e}{e_{r}})^{2}}
\end{equation}
where $\beta$ determines the shape of the Gaussian like ramp-up function, $\lambda_{max}$ refers to maximum value of $\lambda$, $e$ is the current epoch and $e_{r}$ is epoch (ramp-up length) at which $\lambda$ attains its maximum value ($\lambda_{max}$). When $\lambda$ is small, supervision loss dominates. As training progresses, $e$ approaches $e_{r}$ which pushes $\lambda$ towards $\lambda_{max}$, allowing consistency loss to take over. Figure \ref{dynamic_balancing_factor} displays the dynamic balancing curves for different $\beta$'s.

\subsubsection{Inference} During testing, given an unlabeled facial image, we use average of predictions from three networks to infer expression label. However, all networks give comparable performance, so any one of three networks can be used to reduce inference time.
\begin{algorithm}

  \KwInput{ Model f with parameters $\theta = \{\theta_{1}, \theta_{2}, \theta_{3}\}$, dataset(D), learning rate($\eta$), total epochs $e_{max}$, maximum lambda value $\lambda_{max}$}
  %\KwOutput{$\theta = \{\theta_{1}, \theta_{2}, \theta_{3}\}$}
  
  Initialize $\{\theta_{1}, \theta_{2}, \theta_{3}\}$ randomly.
  
   \For{$e = 1,2,..,e_{max}$} 
   {
    \hspace*{1em}Sample mini-batch $D_{n}$ from D\;
    \hspace*{1em}Compute $ p_{i} = f(x,\theta_{i}) \forall x \in D_{n}, (1\leq i\leq3) $\;
    \hspace*{1em}Compute dynamic balancing factor  $\lambda$ using Eq.\eqref{rampup}\;
    \hspace*{1em}Compute joint loss $L$ using Eqs. \eqref{loss}, \eqref{sup_loss} and \eqref{cons_loss}\;
    \hspace*{1em}Update $\theta = \theta - \eta \nabla L$
    
   }      
   return $\{\theta_{1}, \theta_{2}, \theta_{3}\}$
\caption{CCT training algorithm}
\label{CCTtraing_algorithm}
\end{algorithm}

%\end{multicols}

%\begin{multicols}{2}
\section{Datasets and Implementation Details}
\subsection{Datasets}
Aff-Wild2 dataset consisting of 539 videos with 2, 595, 572 frames with 431 subjects, 265 of which are male and 166 female, is annotated for 7 basic facial expressions. Eight of videos have displayed two subjects both of which have been labelled as left and right. Aff-Wild2 is split into three subsets: training, validation and test subsets consisting of 253, 71 and 223 videos respectively. Cropped and aligned frames for all them are made available as a part of the challenge. Since the dataset is highly imbalanced in terms of number of images per expression category, so we use other in the wild datasets like  AffectNet\cite{12}, RAFDB \cite{li2019reliable_45,li2017reliable_46} for pre-training. \textbf{AffectNet} \cite{12} is the largest facial expression dataset with around 0.4 million images manually labeled for the presence of eight (neutral, happy, angry, sad, fear, surprise, disgust, contempt). Images except contempt category are considered for training the model. \textbf{RAF-DB} contains 29672 facial images tagged with basic or compound expressions by 40 independent taggers. Only 12271 images with basic emotion from training set are used.

\subsection{Implementation details}
The proposed work is implemented in PyTorch\footnote{https://pytorch.org} using a three GeForce RTX 2080 Ti GPUs with 11GB memory each. The individual networks in CCT are ResNet-18 pre-trained on large scale face dataset MS-Celeb-1M \cite{msceleb}.  All images in AffectNet, RAFDB are aligned using MTCNN\footnote{https://github.com/ZhaoJ9014/face.evoLVe.PyTorch} \cite{zhang2016_56} and then resized to 224x224. For Aff-Wild2, cropped images provided by organizers are used after resizing them to 224x224. Batch size  is set to 256. Optimizer used is Adam. Learning rate (lr) is initialized as 0.001 for base networks and 0.01 for the classification layer. Further, lr is decayed exponentially by a factor of 0.95 every epoch. Data augmentation includes random horizontal flipping, random erasing and color jitter. $\lambda_{max}$ is set to 0.9 and  $\beta$ to $4.0 $ based on the ablation study in Sec \ref{ablation_dynamic_balancing_factor}.  

\subsection{Evaluation metric}
Evaluation metric used in the challenge evaluation is weighed average of accuracy (33\%) and $F_{1}$ score (67\%). 
Accuracy is defined as fractions of predictions that are correctly identified. It can be written as: 
\begin{equation}\label{eq:6}
Accuracy(Acc) = \frac{Number \hspace{.3em} of\hspace{.3em} Correct\hspace{.3em}Predictions}{Total \hspace{.3em}number \hspace{.3em}of\hspace{.3em} Predictions}
\end{equation}

$F_{1}$ score is defined as weighted average of precision (i.e. Number of positive class images correctly identified out of positive predicted) and recall (i.e. Number of positive class images correctly identified out of true positive class). It can be written as:
\begin{equation}\label{eq:7}
F_{1} \hspace{.3em} score = \frac{ 2 \times  precision \times recall }{ precision + recall}
\end{equation}
And the overall score considered is:
\begin{equation}\label{eq:8}
Overall \hspace{.3em} score = 0.67 \times F_{1} + 0.33 \times Acc
\end{equation}

\section{Results and Discussions}\label{Resultsanddiscussion}
\subsection{Performance Comparison with state-of-the-art methods}
We report our results on the official validation set of expression track from the ABAW 2021 Challenge \cite{ABAW2} in Table \ref{tab:Tab1} . Our best performance achieves overall score of 0.48 on validation set which is a significant improvement over baseline. For the competition, we have submitted multiple models based i) trained on only train set of Aff-Wild2, ii) trained on both train and validation set of Aff-Wild2 and iii) pretrained on AffectNet and finetuned on Aff-Wild2.

\begin{table}[hbt!]

\centering
    \caption{Performance comparison on Aff-Wild2 validation set}
    %\begin{tabular}{p{0.1\textwidth}|p{0.08\textwidth}|p{0.08\textwidth}|p{0.08\textwidth}}
    \begin{tabular}{c|c|c|c}
         \hline
         Method & F1 score & Accuracy &  Overall \\
         \hline
         \hline
         Baseline \cite{ABAW2} & 0.30 & 0.50 & 0.366 \\
         CCT \cite{gera_cct} & \textbf{0.4040} & \textbf{0.6378} & \textbf{0.4814} \\
         \hline
    \end{tabular}
    \label{tab:Tab1}
    %\end{center}
    
\end{table}

\begin{table}
 \centering
    \caption{Performance comparison on Aff-Wild2 Test set}
    \begin{tabular}{c|c|c|c}
    %\begin{tabular}{p{0.1\textwidth}|p{0.08\textwidth}|p{0.08\textwidth}|p{0.08\textwidth}}
         \hline
          Method & F1 score & Accuracy &  Overall \\
         \hline
         \hline
          Baseline \cite{ABAW2} & 0.26 & 0.46 & 0.326 \\
         %Keegs\cite{Keegs} & 0.2545 & 0.507 & 0.3378 \\
         %HUST AUTO1102\cite{HUSTAUTO1102} & 0.2809 & 0.5822 & 0.3803 \\
         Kawakarpo \cite{kawakarpo} & 0.29 & 0.6491 & 0.4082 \\
         %SZTU-CityU\cite{SZTU} & 0.3073 & 0.6234 & 0.4116 \\
         NTUA-CVSP \cite{NTUACVSP} & 0.3367 & 0.6418 & 0.4374 \\
         Morphoboid \cite{Morphoboid} & 0.3511 & 0.668 & 0.4556 \\
         FLAB2021 \cite{FLAB2021} & 0.4079 & 0.6729 & 0.4953 \\
         %NISL-2021\cite{NISL2021} & 0.4311 & 0.6538 & 0.5046 \\
         STAR \cite{STAR} & 0.4759 & 0.7321 & 0.5604 \\
         Maybe Next Time \cite{Maybe} & 0.6046 & 0.7289 & 0.6456 \\
         CPIC-DIR2021 \cite{CPIC}& 0.6834 & 0.7709 & 0.7123 \\
        Netease Fuxi Virtual Human \cite{Netease} & \textbf{0.763} & \textbf{0.8059} & \textbf{0.7777} \\
         \hline
         Ours\cite{gera_cct} & 0.361 & 0.675 & 0.4646 \\
         \hline
    \end{tabular}
%\end{center}
    \label{tab:Tab2}
\end{table}

Table \ref{tab:Tab2} presents the performance comparison on test set w.r.t other participating teams\footnote{https://ibug.doc.ic.ac.uk/resources/iccv-2021-2nd-abaw/}. Our top performing model was trained on whole train set of Aff-Wild2 using proposed model which is pretrained on AffectNet. Clearly, out model gives superior performance compared to many of teams. Since, we have trained model for single task and not used any of audio and video features, so performance is not as good as teams using multi-task learning with video features.

\begin{table}[hbt!]

 \centering
    \caption{Performance comparison w.r.t different number of models in CCT on validation set}
    \begin{tabular}{c|c|c|c|c}
    %\begin{tabular}{p{0.07\textwidth}|p{0.08\textwidth}|p{0.08\textwidth}|p{0.08\textwidth}}
         \hline
         No of models & Pretrained & F1 score & Accuracy &  Overall \\
         \hline
         \hline
         1 & AffectNet & 0.3311 & 0.5705 & 0.41 \\
         2 & AffectNet & 0.3744 & 0.621 & 0.456 \\
         3 & AffectNet & \textbf{0.4040} & \textbf{0.6378} & \textbf{0.4814} \\%\textbf{0.3865} & \textbf{0.63} & \textbf{0.467} \\
         4 & AffectNet & 0.366 & 0.613 & 0.447 \\
         \hline
 \end{tabular}
    %\end{center}
\label{tab:tab_no_of_models}    
\end{table}

\begin{table}[hbt!]

 \centering
    \caption{Performance comparison w.r.t different dataset for pretraining in CCT on validation set.}
    \begin{tabular}{c|c|c|c}
    %\begin{tabular}{p{0.07\textwidth}|p{0.08\textwidth}|p{0.08\textwidth}|p{0.08\textwidth}}
         \hline
         Pretrained dataset& F1 score & Accuracy &  Overall \\
         \hline
         \hline
          MS-Celeb-1M & 0.378 & 0.609 & 0.454 \\
          RAFDB & 0.383 & 0.615 & 0.460 \\
          AffectNet & \textbf{0.4040} & \textbf{0.6378} & \textbf{0.4814} \\%\textbf{0.3865} & \textbf{0.63} & \textbf{0.467} \\
         \hline
 \end{tabular}
%    \end{center}
    \label{tab:tab_pretrained_dataset}
\end{table}

\begin{table}[hbt!]
 \centering
    \caption{Performance comparison with and without oversampling in CCT on validation set using MS-Celeb-1M pertrained model.}
    \begin{tabular}{c|c|c|c}
    %\begin{tabular}{p{0.07\textwidth}|p{0.08\textwidth}|p{0.08\textwidth}|p{0.08\textwidth}}
         \hline
         Oversampling & F1 score & Accuracy &  Overall \\
         \hline
         \hline
          Without &  0.378 & 0.609 & 0.454 \\
          With & 0.357 & 0.58 & 0.43 \\
          
         \hline
 \end{tabular}
    
\label{tab:tab_oversampling}    
\end{table}

\begin{table}[hbt!]
    \centering
    \caption{Impact of consistency loss}
    \begin{tabular}{c|c|c|c}
         \hline
        Loss & F1 score & Accuracy &  Overall \\
         \hline
         \hline
         CE Loss &  0.3634 & 0.610 & 0.445 \\
         CE Loss + Consistency Loss &  0.378 & 0.609 & 0.454 \\
         \hline
    \end{tabular}
    
    \label{tab:consistency_loss_vs_celoss}
\end{table}

\begin{table}[hbt!]
    \centering
    \caption{Impact of dynamic balancing factor ($\lambda$) for different values of $\beta$ in Eq. \ref{rampup} and Fig. \ref{dynamic_balancing_factor} using AffectNet pertrained model. }
    \begin{tabular}{c|c|c|c}
         \hline
        $\beta$ & F1 score & Accuracy &  Overall \\
         \hline
         \hline
         0.1 &  0.352 & 0.593 & 0.432 \\
         0.65 &  0.378 & 0.609 & 0.454 \\
         1.0 &  0.390 & 0.6378 & 0.472 \\
          1.5 &  0.378 & 0.6269 & 0.460 \\
          2.0 &  0.3969 & 0.6160 & 0.469 \\
          3.0 &  0.4051 & 0.6226 & 0.4769 \\
          4.0 &  \textbf{0.4040} & \textbf{0.6378} & \textbf{0.4814} \\
          5.0 &  0.3931 & 0.6387 & 0.4742 \\
          %3.0 &  0.390 & 0.638 & 0.472 \\
           %4.0 &  0.390 & 0.638 & 0.472 \\
         \hline
    \end{tabular}
    
    \label{tab:dynamic_balancing_factor}
\end{table}

\subsection{Ablation studies}

\subsubsection{Influence of number of networks:}
Table \ref{tab:tab_no_of_models} shows the influence of number of networks that are collaboratively trained in CCT. It can be observed that model with 3 networks performs the best in the presence of noise. This is because of the region where the 3 networks come into consensus is relatively smaller, thereby avoiding more noisy labels during training. However, with 4 networks, the number of parameters also significantly increases and building consensus is quite difficult. 

\subsubsection{Impact of dynamic balancing factor} \label{ablation_dynamic_balancing_factor}
Since, as per Eq. \ref{rampup}, if the best $\beta$ is determined, automatically the best $\lambda$ gets fixed. So, we determine the best $\beta$ by computing performance for different values of beta in Table \ref{tab:dynamic_balancing_factor}. $\beta = 4$ gave best performance in Aff-Wild2 validation set.

\subsubsection{Effect of pretrained dataset}
We compared the influence of different datasets for pretrained model in Table \ref{tab:tab_pretrained_dataset}. Clearly, AffectNet dataset due to its large scale gives better performance compared to RAFDB.

\subsubsection{Influence of oversampling}
Table \ref{tab:tab_oversampling} shows that oversampling is not effective for training. So, all test submission are based on training without oversampling.

\subsubsection{Impact of consistency loss}
Table \ref{tab:consistency_loss_vs_celoss} evaluates the effect of consistency loss  w.r.t training using only supervision loss. Clearly, dynamic transition using combined loss is more effective than training using only supervision loss.
\section{Conclusions}
In this paper, we present robust framework called CCT for effectively training a FER system with noisy annotations on Aff-Wild2 dataset. CCT combated the noisy labels by co-training three networks. Supervision loss at early stage of training and gradual transition to consistency loss at later part of training ensured that the noisy labels did not influence the training. Both the losses are balanced dynamically. Our results on challenging dataset as a part of ABAW 2021 competition demonstrate the superiority of our method as compared to many others methods presented without using any audio and landmarks information. In the future work, we would like to test our model for valence-arousal estimation and facial action unit recognition tasks.

\section{Acknowledgments}
We dedicate this work to Our Guru Bhagawan Sri Sathya Sai Baba, Divine Founder Chancellor of Sri Sathya Sai Institute of Higher Learning, PrasanthiNilyam, A.P., India. We are also grateful to D. Kollias for all patience and support.

\bibliographystyle{unsrt}  
%\bibliography{references}  %%% Remove comment to use the external .bib file (using bibtex).
%%% and comment out the ``thebibliography'' section.
\bibliography{references}

%%% Comment out this section when you \bibliography{references} is enabled.
%\end{multicols}
\end{document}